\documentclass[10pt, a4paper]{article}
\usepackage{tipa}
\usepackage[final]{lrec2026} 
\usepackage{adjustbox}
\usepackage{todonotes}
\usepackage{lingmacros}
\usepackage{cleveref}
\usepackage{subcaption}
\usepackage{multirow}
 \usepackage{booktabs}
\title{Saar-Voice: A Multi-Speaker Saarbrücken Dialect Speech Corpus}

\name{Lena S. Oberkircher, Jesujoba O. Alabi, Dietrich Klakow, Jürgen Trouvain } 

\address{Language Science and Technology (LST), Saarland University, \\
         Saarbrücken, Germany \\
         \{s8leober@stud, jalabi@lsv, dietrich.klakow@lsv, trouvain@lst\}.uni-saarland.de\\}

\abstract{
Natural language processing (NLP) and speech technologies have made significant progress in recent years; however, they remain largely focused on standardized language varieties. Dialects, despite their cultural significance and widespread use, are underrepresented in linguistic resources and computational models, resulting in performance disparities. To address this gap, we introduce \textbf{Saar-Voice}, a six-hour speech corpus for the Saarbrücken dialect of German. The dataset was created by first collecting text through digitized books and locally sourced materials. A subset of this text was recorded by nine speakers, and we conducted analyses on both the textual and speech components to assess the dataset's characteristics and quality.
We discuss methodological challenges related to orthographic and speaker variation, and explore grapheme-to-phoneme (G2P) conversion. The resulting corpus provides aligned textual and audio representations. This serves as a foundation for future research on dialect-aware text-to-speech (TTS), particularly in low-resource scenarios, including zero-shot and few-shot model adaptation. 
 \\ \newline  }

\begin{document}

\maketitleabstract

\section{Introduction and Background}

The field of natural language processing (NLP) has seen substantial advances in recent years, driven by large-scale models and increased computational resources. Nevertheless, research and tools remain heavily focused on standardized language varieties, leaving dialectal varieties underrepresented despite their widespread use and importance to speakers' cultural identity~\citeplanguageresource{blasi-etal-2022-systematic}. In Germany, for example, more than 40\% of the population regularly speaks regional dialects, yet these varieties are often subject to social stereotyping~\citep{adler2022dialekt}. This marginalization is reflected in the technological landscape: large language models (LLMs) and speech processing systems frequently struggle with dialectal variation, as they are predominantly trained on standardized language data~\citep{kruckl-etal-2025-improving}. Consequently, recent years have seen growing efforts to develop language resources for dialectal varieties~\citeplanguageresource{blaschke-etal-2024-maibaam,faisal-etal-2024-dialectbench,blaschke25_interspeech} and to systematically evaluate the performance of existing NLP systems on such data~\citep{bui-etal-2025-large,munoz-ortiz-etal-2025-evaluating}.

In this paper, we contribute to this line of work by focusing on the Saarbrücken dialect of German spoken in the state of Saarland through the creation of \textbf{Saar-Voice}, a six-hour multi-speaker speech corpus.\footnote{The dataset is available on \url{https://huggingface.co/datasets/UdS-LSV/Saar-Voice}.} In the following subsections, we provide some basic background about the Saarbrücken dialect, including the main linguistic characteristics of the dialect.

\subsection{Dialect areas}
The Saarland is one of Germany's 16 federal states (Bundesländer), located in the southwest of the country and bordering France and Luxembourg. The Saarland has around 1 million inhabitants. The state is, much like the rest of Germany, a linguistically diverse region, with many variants of the standard language, from regiolects to dialects.

\begin{figure}
    \centering
    \includegraphics[width=1\linewidth]{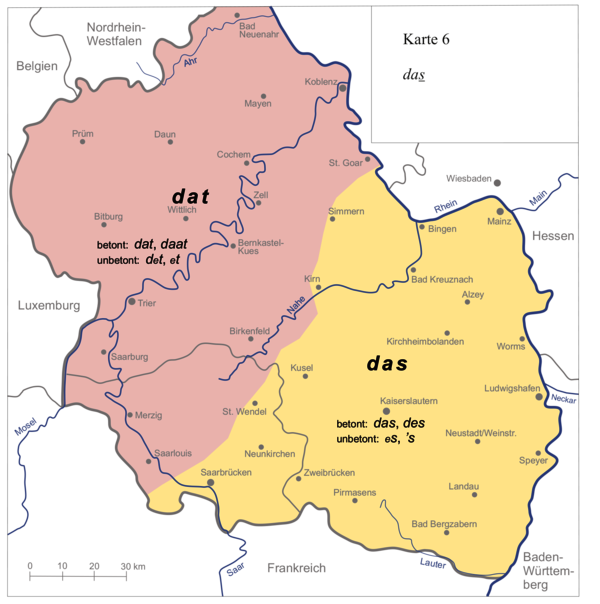}
    \caption{The isogloss \textit{dat-das-line} distinguishing areas in which Rhine Franconian dialects (“das”) and Moselle Franconian dialects (“dat”) are spoken \citep{drenda2008} in Saarland (smaller area in the south-west marked with thin border lines) and Rhineland-Palatinate.}
    \label{fig:dat-das}
\end{figure}

A typical problem in linguistics is defining which exact dialect is meant with a given name. In the region of the Saarland, the broad range of regiolects and dialects are loosely referred to as “Saarländisch”, but this is not a precise term. On a large scale, “Saarländisch” consists of two different dialect areas: Rhine Franconian and Moselle Franconian (see Figure \ref{fig:dat-das}). Both dialect areas are rather diverse on all linguistic levels, mainly phonology, morpho-syntax and lexicon. Looking more closely, there is large variance even within the two regions.

We decided in this paper to consider the Rhine Franconian dialect spoken in and around the capital of and largest city in the Saarland, Saarbrücken, for which a grammar \citeplanguageresource{steitz1981saarbr} and a suggested dictionary \citeplanguageresource{braun1984saarbrucker} exist, both authored by linguists, as well as a community of active authors (with no background in linguistics).

\subsection{Spelling}
One big advantage of a standard language is the existence of a standard spelling. For (German) dialects standardized orthographies are missing, which leads to high variance in spelling between and even within speakers. Often, speakers use spontaneous spelling, orthographically representing the phonology each word as they best see fit. For instance, the exact spelling of “es gebbd” (e. ‘there is’, ‘there are’) can vary a lot: “es gebbd, gebd, gäbbd, gäbd, gebbt, gäbt”; or “Zeidung” (e. “newspaper’) as “Zeidung, Tseidung, Dsaidung”.

Generally, speakers use the letters available in the Standard German alphabet (“a-z, ä, ö, ü” and “ß”). Additionally, depending on the speaker, certain diacritics that are not part of the standard language may be used. For example, “ò” may be used to represent the phoneme \textit{/\textopeno/} (such as in “dò”, “dòò”, “do” or ”doo”, all representing /\textipa{d\textopeno\textlengthmark}/, meaning ”there”), and in some cases \textit{á} to represent the phoneme \textit{/\textturna/} as in “vákaafe” (/\textipa{f\textturna ka\textlengthmark f\textschwa}/, e. 'sell').


\begin{table*}[ht]
    \centering
    \begin{tabular}{l|cccc}
        \toprule
        \textbf{Source} &  \textbf{Domain} & \textbf{\# Sent} & \textbf{\# Words} & \textbf{\# U. Words} \\
        \midrule
    \multicolumn{5}{l}{\textbf{Printed Books}} \\
        An da Saar gefonn  & Poetry& 698& 6,228& 1,978\\
        Geschaffd - Gelääbd& Poetry; Prose& 394& 3,302& 1,214\\
        Saa, was de willschd& Prose; Poetry& 2,216& 19,711& 3,147\\
        Was wääs dann isch& Poetry; Prose& 2,525& 18,445& 3,490\\
    \multicolumn{5}{l}{\textbf{Locally Sourced}} \\
        (Locally Sourced Texts)& Prose; Folktale; Poetry& 2,838& 26,789& 5,221\\
    \multicolumn{5}{l}{\textbf{Localized Translation}} \\
        MASSIVE German & Localized Data& 101& 805& 394\\
         \midrule
         \multicolumn{2}{l}{\textbf{Total}} & 8,772& 75,280& 11,303\\
        \bottomrule
    \end{tabular}
        \caption{Textual corpus statistics of the collected texts for the Saarbrücken dialect.}
    \label{tab:text_stats}
\end{table*}

\subsection{Differences between dialect and standard language}
Were dialects simply accents of the standard language, the approach to creating any speech system would be to simply 'translate' the standard pronunciation to the regional pronunciation (ideally with the intonation of the accent). However, dialects are not just the standard language with regional accents. In fact, they often differ to written standard language on various levels:

\begin{itemize}
    \item \textbf{Orthography and phonology:} Dialectal writing reflects phonological distinctions that are not represented in Standard German orthography. For example, vowel quality and length may be encoded differently, and consonant correspondences do not always align with standard spelling conventions.

    \item \textbf{Morpho-syntax:} Grammatical markers such as case or gender markings may differ between the dialect and standard language, among other morpho-syntactic differences.
    \begin{itemize}
        \item Standard German: \textit{ein} (indefinite article, masc./neut.) + \textit{schön} (nice) + \textit{Überraschung} (surprise; fem.) $\rightarrow$ \textit{eine schön\underline{e} Überraschung}
        \item Dialect: \textit{e} (indefinite article, generalized form) + \textit{scheen} (nice) + \textit{Iwwerraschung} (surprise) $\rightarrow$ \textit{e scheen\underline{i} Iwwerraschung}
    \end{itemize}
    Here, both the article system and adjectival inflection differ from Standard German.

    \item \textbf{Lexical variation:} Dialects employ vocabulary items that are absent from Standard German or derived from different etymological sources, e.g., \textit{Gellerieb} (‘carrot’, from “gelbe Rübe”, in Standard German “Karotte” or “Möhre”), \textit{Grumbier} (‘potato’, from “Grundbirne”, in Standard German “Kartoffel”), or \textit{Troddwaa} (‘sidewalk’, from French “trottoir”, in Standard German “Gehsteig” or “Bürgersteig”).
\end{itemize}

An important feature of dialect texts is that they are rarely found in the everyday 'linguistic landscape'. This has not only consequences for feelings of insecurity of spelling among speakers, including inconsistencies of spelling of identical words in the same text, but also for potential readers.
Most \textit{speakers} of dialects are not very good \textit{readers} of dialect text, though they are very fluent and competent readers of standard language (but maybe speakers with a regional accent). 

\subsection{Text sources}
There are different sources of dialect-writing, for example:
\begin{itemize}
    \item web-based: chats, blogs
    \item books written by (non-professional) dialect authors
    \item research pieces, e.g. dictionaries, monographs and translations of local dialects
    \item texts from authors organized in dialect clubs
    \item public media, e.g. news in local radio stations, glosses in local newspaper
\end{itemize}

Depending on selected types of text, the styles can hugely differ. For instance, many dialect writers produce poems, in which the stanzas often rhyme at the end. In contrast, in chats reduced forms of written conversations can be observed which are again quite different to some dialect prose.
Thus, the selected text source used for later steps can have a strong bias of style. This is especially important when creating a speech corpus, which may in the end contain biases in pronunciation and intonation.

\section{Related Work}
\paragraph{German Speech Resources} Several resources exist for German spoken language research, but most are limited in dialectal coverage. Existing datasets include unannotated speech collections for speech representation learning, such as VoxLingua107~\citeplanguageresource{Valk2020VOXLINGUA107AD}, VoxPopuli~\citeplanguageresource{wang-etal-2021-voxpopuli}, JesusDramas, WikiTongues, and MMS ULAB v2~\citeplanguageresource{chen-etal-2024-towards-robust}, which are mostly derived from social media, religious texts, or read speech for Standard German. Automatic speech recognition (ASR) datasets include FLEURS~\citeplanguageresource{Conneau2022FLEURSFL}, and dataset from initiatives like Mozilla Common Voice provide more structured recordings, while TTS datasets such as MLS~\citeplanguageresource{pratap20_interspeech} (also useful for ASR), FLEURS-R~\citeplanguageresource{ma24c_interspeech}, Thorsten-Voice Dataset~\citeplanguageresource{thorsten_muller_2024}, and CML-TTS~\citeplanguageresource{Cmltts2023}, created from MLS, include high-quality recordings of standard German.

In recent years, some efforts have targeted German dialects for ASR and TTS, covering varieties such as Swiss German~\citeplanguageresource{Plss2020SwissPC}, Viennese~\citeplanguageresource{schabus2013joint,pucher-etal-2010-resources}, non-Vienna Austrian~\citeplanguageresource{Pucher2017}, Luxembourgish~\citeplanguageresource{Steiner2017}, and three dialect groups spoken in Southeast Germany (Franconian, Bavarian, Alemannic)~\citeplanguageresource{blaschke25_interspeech}. However, several dialects, such as the Saarbrücken dialect, remain underrepresented, and to the best of our knowledge only a small corpus is available in Bible MMS~\citeplanguageresource{lux24_interspeech}. Furthermore, a recent survey~\citep{blaschke-etal-2024-dialect} showed that German speakers are interested in language technologies that can process dialectal (audio) input, and creating and evaluating such systems requires developing speech corpora for underrepresented dialects.

To address these gaps, we introduce \textbf{Saar-Voice}, a multi-speaker speech corpus for the Saarbrücken dialect, providing high-quality recordings suitable for training and evaluating modern TTS systems.

\paragraph{Speech Corpora for Dialects and Low-Resource Languages} Beyond German-specific efforts, the creation of speech corpora for dialectal and low-resource settings has been widely explored across languages~\citep{Zampieri_Nakov_Scherrer_2020,ramponi-2024-language,lent-etal-2022-creole,GUELLIL2021497,alabi-etal-2025-charting}. Studies on dialectal variants across a range of languages, such as Arabic~\citep{malmasi-zampieri-2016-arabic,djanibekov-etal-2025-dialectal,talafha-etal-2025-nadi}, English~\citep{ahamad-etal-2020-accentdb,xiao-etal-2023-task,olatunji-etal-2023-afrispeech}, and African languages~\citep{ahia-etal-2024-voices,emezue-etal-2024-igboapi}, highlight the importance of accounting for regional and social variation when developing speech datasets, as models trained on standardized varieties often fail to generalize to non-standard speech~\citep{diab-2016-processing,aji-etal-2022-one,ahia-etal-2024-voices,alabi25_interspeech}.

In parallel, a substantial body of research has focused on speech data collection for low-resource and underrepresented languages. These settings are typically characterized by limited availability of speakers, scarce linguistic resources, and, in some cases, the absence of standardized orthography~\citep{blaschke-etal-2024-dialect}. As a result, corpus construction in such contexts often relies on adaptable methodologies, including crowd-sourced data collection, community-driven recording initiatives, and semi-supervised or lightweight annotation strategies~\citep{emezue-etal-2024-igboapi,olatunji-etal-2023-afrispeech}.

Despite progress, many speech datasets focus only on common languages and pay little attention to dialect differences. In addition, datasets for low-resource languages often lack consistent annotation or enough speakers, which limits their usefulness for training and evaluating models. To address these challenges, we introduce a carefully curated speech corpus of a German dialect, with nine speakers and about six hours of recordings. The dataset is designed to provide a compact but high-quality resource for studying non-standard German speech and for training and benchmarking speech models, especially TTS, in low-resource dialect settings.

\begin{table}
    \centering
    \small
    \begin{tabular}{l c}
    \toprule
    \textbf{Speakers} &  \\
    Total speakers & 9 \\
    Gender ratio & 4F / 5M \\[3pt]
    \midrule
    \textbf{Speech} &  \\    
    Recorded sentences & 4,871 \\
    Total duration (hh:mm:ss) & 05:54:48 \\
    Avg. duration per sent. (sec) & 4.37 \\
    Min duration of sent. (sec) & 1.59 \\
    Max duration of sent. (sec) & 14.02 \\
    \bottomrule
    \end{tabular}
    \caption{Saar-Voice Corpus Statistics.}
    \label{tab:corpus_stats}
\end{table}

\section{Corpus Design}

This section describes the methodology used to create the Saar-Voice corpus.

\subsection{Text Collection}

\paragraph{Digitization of Printed Books} For this study, we digitalized four books available in print at the Saarland University library. The books are “An da Saar gefonn: volkstümliche Gedichte in Saarbrücker Mundart” \citeplanguageresource{jungmann1993andasaar}, “Geschaffd - Gelääbd: Mundarttexte” \citeplanguageresource{fox1994geschaffd}, “Saa, was de willschd: Mundart-Kolumnen” \citeplanguageresource{fox2012saawasde} and “Was wääs dann isch...?!” \citeplanguageresource{eckert1995waswaas}. These texts were scanned and digitized using an online OCR software\footnote{https://ocr.ac/de}. This OCR software was chosen in particular as it was one of the few softwares that had close to no issue recognizing the special characters “á” and “ò”, which caused great problems with other softwares and were usually misclassified as “ä” and “ö”, respectively. Remaining errors introduced by the OCR process were removed through manual review by a native speaker of the dialect, while also cross-referencing the original source texts. 5,833 sentences were collected this way, amounting for 66.6\% of the full dataset. 

\paragraph{Locally Sourced Texts} 2,838 sentences, accounting for 32.4\% of the total data, were collected from internally available texts written by authors from the local community. These texts were already available digitally and only checked for possible spelling errors by a native speaker.

\paragraph{Localized Translations} Lastly, we sampled 101 German sentences from the MASSIVE dataset ~\citeplanguageresource{fitzgerald-etal-2023-massive}. These sentences were manually translated with the support of the dictionary of  \citetlanguageresource{braun1984saarbrucker}. After translation, the data was localized by replacing entities with localized entities, such as locations and personal names. These localized entities were manually inserted to further emphasize dialect-specific variation. Localized location names were taken at random from \citetlanguageresource{brown1981necknamen}, as in Example~\ref{ex:localized-example}. Additionally, numbers were spelled out in dialect orthography rather than given as numerals, as in Example~\ref{ex:localized-example_numbers}, to ensure consistent pronunciation across speakers.

\enumsentence{\label{ex:localized-example}\small\textbf{German Original}: Bitte plane ein Treffen mit \underline{Petra} in \underline{Wiesbaden} am Mittag.\\
\textbf{German Original (Entities Replaced)}: Bitte plane ein Treffen mit \underline{Anna} in \underline{Oberbexbach} am Mittag.\\
\textbf{Translated}: Bidde plaan e Dreffe midd \underline{Anna} in \underline{Oberbexbach} am Middaach.\\
\textbf{Localized}: Bidde plaan e Dreffe midd \underline{Anna} in \underline{Owwerbeddschbach} am Middaach.
}
\enumsentence{\label{ex:localized-example_numbers}\small\textbf{German Original}: Erinnere mich an \underline{Laura}s Geburtstag \underline{24} Stunden vorher.\\
\textbf{German Original (Entities Replaced)}: Erinnere mich an \underline{Melanie}s Geburtstag \underline{19} Stunden vorher.\\
\textbf{Translated}: Erinner misch an \underline{Melanie}s Geburdsdaa \underline{19} Schdunne vòòrhäär.\\
\textbf{Localized}: Erinner misch an \underline{Melanie}s Geburdsdaa \underline{neindsehn} Schdunne vòòrhäär.
}

An overview over text statistics for each of the collected resources can be found in Table \ref{tab:text_stats}. This includes an estimation of the domains each of the resources cover, as annotated by a native speaker. Most resources are collections of texts rather than a single running texts. Therefore, for resources that are associated with multiple domains, domains are reported in descending order of frequency.

\subsection{Speaker Recruitment}
A total of nine participants were recruited using convenience sampling from people known to the research team. The speaker group consisted of four female and five male speakers. Age of participants was collected in categorical ranges. The largest age group was 26-30 years (n=3), followed by 31-35 years (n=2). The remaining participants were distributed across the ranges 18-25, 51-55, 56-60 and 61-65 (n=1 each).

Participants were either native speakers of a Rhine Franconian dialect, or native speakers of a closely related regional dialect with high familiarity and regular, long-term exposure to the target dialect. Self-identified dialect labels included: Saarländisch / Saarländischer Dialekt (\textit{Saarland Dialect}; n=5), Moselfränkisch (\textit{Moselle Franconian}, n=1), Rheinfränkisch (\textit{Rhine Franconian} n=1), Rhein-Moselfränkischer Grenzdialekt (\textit{Rhine-Moselle Franconian Border Dialect}, n=1), and Platt  (\textit{local term for the dialect}, n=1).

All speakers reported speaking the dialect daily (n=6) or at least multiple times a week (n=3) in the present time, and the majority (n=7) reports having spoken mainly the dialect during their childhood.

All speakers are also (native) speakers of Standard German and were “alphabetized” with Standard German spelling.

\begin{table*}[t]
\scalebox{0.9}{
    \centering
    \small
    \begin{tabular}{l c c cc cc c}
    \hline
    \textbf{Speaker} & \textbf{Gender} & \textbf{Total audio (hh:mm:ss)} &\textbf{\# sentences}& \textbf{Mean F0} &  \textbf{Median F0}&\textbf{F0 range (5–95\%)}& \textbf{W/s}\\
    \hline
        P01& F& 00:23:17 &301& 203&  200&141-278& 1.8249\\
        P02& F& 01:02:28 &1075& 251&  249&203-312& 2.4166\\
        P03& M& 01:15:37 &899& 115&  107&80-176&1.7352\\
        P04& M& 00:27:56 &400& 143&  140&100-196&2.2056\\
        P05& F& 00:27:18 &399& 213&  212&175-256&2.0203\\
        P06& M& 00:18:43 &300& 119&  115&84-160&2.3334\\
        P07& M& 00:53:21 &597& 171&  164&102-258&1.5016\\
        P08& F& 00:44:45 &600& 201&  196&156-265&1.7920\\
        P09& M& 00:21:22 &300& 147&  141&101-203&1.8053\\
    \hline
    \end{tabular}
    }
    \caption{Speaker-level acoustic statistics. F0 in Hz. W/s = words per second.}
    \label{tab:speaker_stats}
\end{table*}
\subsection{Recording Setup}
Participants were invited to the departmental soundproof recording booth for audio recordings. 
The texts were presented to the speakers sentence by sentence on a
21.5-inch Full HD monitor. Audio was recorded at a sampling rate of 44.1kHz using a \textit{DAP 2011} microphone, a high-quality directional microphone, using the SpeechRecorder Software \citep{draxler-jansch-2004-speechrecorder}. Each recording session took around 1 hour, during which, depending on the participant, 200-300 sentences were recorded in batches of 100.

The speakers controlled their own pace of sentence presentation and recording by mouse clicks on a specified icon. If the speaker felt unsure or had a slip of the tongue s/he had the option for a new recording. Also, the recording supervisor (the first author) had the possibility to ask the speaker to restart the recording of a sentence.

Sentences presented to the speakers were not sampled independently, but derived from a longer, coherent text corpus. The entire textual corpus was concatenated, segmented into sentences and then grouped into chunks of 100 consecutive sentences. For each session, such chunks were randomly selected. This semi-randomized procedure ensured topical and lexical continuity within one session, while still maintaining coverage across the corpus and across domains.

Recording continuous passages like this offers several advantages. Besides keeping the speaker engaged in a cohesive story, it supports a more natural prosody and fluency, leading to consistent pronunciation of recurring and possibly unfamiliar lexical items, and enabling the resulting book subset to be used for long-context TTS experiments.

\section{Saar-Voice Corpus}
This section presents an overview of the created corpus.
\subsection{Data Statistics}



Table \ref{tab:corpus_stats} summarizes the dataset statistics of Saar-Voice. The dataset consists of approximately six hours of speech, comprising 4,871 utterances from nine participants. The average length of the recorded 
sentence is about four seconds, with minimum and maximum durations of two and 14 seconds, respectively; these are moderately short but of good quality.

\subsection{Phoneme Coverage}
We applied Epitran~\citep{Mortensen-et-al:2018} a multilingual grapheme-to-phoneme (G2P) model on the recorded text in Saar-Voice. Our analysis shows that the dataset contains 38 distinct phonemes, which lies in the typical range of phonemes estimated for Standard German \cite{kohler1990german} and also the Saarbrücken dialect \citeplanguageresource{steitz1981saarbr} and \citeplanguageresource{braun1984saarbrucker}, if the vowels of German loan words are included. The most common phoneme is /\textipa{d}/, likely due to the voicing of many occurrences of consonants in the dialect that are typically unvoiced in Standard German (e.g., /\textipa{t}/ $\rightarrow$ /\textipa{d}/). It occurs 13,963 times within the speech corpus, and is followed by /\textipa{n}/ (11,284 occurrences), /\textipa{s}/ (10,963), /\textipa{\textscr}/ and /\textipa{e}/ (both 10,547 occurrences).

The least common phoneme is /\textipa{\o}/, occurring only 4 times in the entire speech corpus, followed directly by /\textipa{\textscoelig}/ (6 occurrences). Both are realizations of the Umlaut “\"o”, which is rather rare in the text itself, appearing only 10 times in total. They are followed by /\textipa{\textscy}/ (13 occurrences), /\textipa{y}/ (26 occurrences) and /\textipa{\textyogh}/ (64 occurrences).
All rounded front vowels /\textipa{\textscoelig}, \textipa{\textscy}, \textipa{y}/ can be regarded as phonemes from Standard German loan words.

The Epitran G2P model is multilingual and designed to cover standard German rather than the Saarbrücken dialect. While generally effective for estimating the phoneme coverage of this dialect corpus, the implications of this mismatch are discussed in \Cref{sec:issues}.








\subsection{Speaker Variability}
Analysis of speaker variability shows that mean F0 (pitch) differs by gender. For male speakers, the mean F0 ranges from 115 Hz to 171 Hz (with a median of 107-164 Hz), while for female speakers it ranges from 201 Hz to 251 Hz (with a median of 196-249 Hz). These statistics indicate moderate inter-speaker variability in both pitch and tempo, reflecting natural differences in speech across male and female voices. We also observe speech rates ranging from 1.5 to 2.4 words per second across both genders, with values evenly distributed within each gender.~\Cref{tab:speaker_stats} provides a detailed breakdown for all speakers.






\subsection{Audio Quality}

\begin{figure}[t]
    \centering
        \centering
        \includegraphics[height=1.5in]{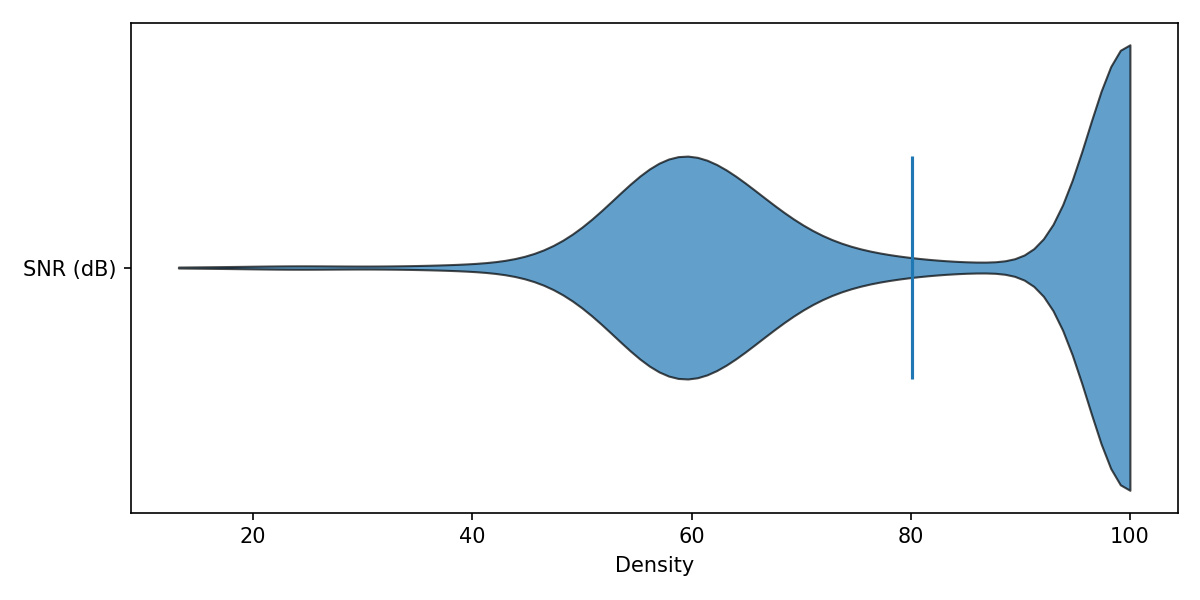}
        \caption{SNR distribution on Saar-Voice.}
        \label{fig:SNR_regular}
    ~ 
    \label{fig:SNR}
\end{figure}

We evaluate the signal-to-noise ratio (SNR) for all the recorded audio samples in Saar-Voice using the WADA-SNR~\citep{kim08e_interspeech} algorithm and plot the distribution in Figure \ref{fig:SNR_regular}. The results show that most samples fall within the clean audio threshold, with SNR values ranging from 20 to 99, confirming the high overall quality of the Saar-Voice dataset.


\section{Issues}
\label{sec:issues}
In this section, we discuss the practical challenges and issues encountered during the corpus creation process.

\subsection{Data-Specific Issues}
A central challenge of building a corpus for a low-resource dialect like the variant of Rhine Franconian discussed in this project is the absence of standardized orthography for the target dialect. Unlike Standard German, the dialects of the Saarland do not have any official spelling conventions or vocabulary lists. Although dialect dictionaries like \citetlanguageresource{braun1984saarbrucker} can serve as guidance, they do not provide an official, normative reference. Orthographic variation is thus unavoidable and inherent to the dialect. It further affects vocabulary size estimates, as orthographic variants of the same lexical item are counted as distinct types.

Ignoring such variation or attempting to impose a standardized orthography would in many cases even go against what native speakers actually desire: \citet{blaschke-etal-2024-dialect} show that 65\% of their participating speakers of German dialects explicitly oppose the introduction of a standardized orthography. 
Our own questionnaire showed no clear consensus regarding standardization. Taken together, these results suggest that there is no stable community-wide preference for standardization, and the more representative survey by \citet{blaschke-etal-2024-dialect} indicates the opposite. Any attempt at standardization would risk privileging one subset of speaker preference over others, potentially misrepresenting actual usage practices, and harming user trust.

The choice of OCR software presented an additional technical challenge. Most available systems, such as Python's \textit{tesseract} library, are optimized and effective for Standard German, but showed a substantial difficulty processing dialect-specific characters such as “á” or “ò”, as well as recognizing entire character strings which are not part of Standard German's inventory. This resulted in recognition errors and, even with the final choice of OCR software, required manual corrections, introducing additional processing effort and potential inconsistencies. 

Further issues with available technologies arise in the lack of dedicated tokenization tools for the dialect. This led to being able to provide only an approximate vocabulary segmentation by whitespace tokenization. Without dialect-specific morphological or tokenization resources, which are hard to create due to data sparseness, word boundary detection may be imperfect. This in turn affects vocabulary size estimation and lexical frequency calculations.

Beyond orthography issues, the composition of the corpus itself raises questions of linguistic representativeness. The genre of poetry makes up over one third (35.9\%) of the dataset. This may, due to the genre's metrical and rhyming conventions, introduce lexical and especially prosodic and suprasegmental patterns that are unlikely to reflect the natural spoken dialect. More broadly, all textual resources, including prose and localized translations, represents the written register language, which inherently diverges from spontaneous speech. Nonetheless, it accurately represents the distribution of written data within this specific dialectal landscape. Additionally, the dataset lacks spontaneous speech. While this is less critical for the dataset's primary intended use case of TTS synthesis, it may limit its application to downstream tasks such as ASR. Future work should therefore focus on expanding the corpus with transcribed spontaneous speech to improve linguistic representativeness and genre coverage, and broaden its utility.

\subsection{Speaker-Specific Issues}
Although all speakers were carefully selected as Rhine Franconian speakers, variation within the speaker pool naturally remains inevitable. A variety of self-assessment data from the provided questionnaire indicates high overall proficiency / fluency, but variance appears in dialect usage patterns. Only one speaker reports speaking predominantly exclusively dialect with little to no Standard German influences, while most described their speech as a mixture of dialect and standard language. Some even indicated that they predominantly speak Standard German. 

These differences, as well as considering small regional differences in the dialect, suggest that the corpus reflects not a uniform dialect realization, but rather a spectrum of usage patterns. This introduces additional variability for downstream modeling tasks.

\subsection{G2P-Specific Issues}

\begin{table*}[t]
\centering
\begin{tabular}{l l l l}
    \hline
    \textbf{Word} & \textbf{English} & 
    \textbf{Epitran} & \textbf{Ours} \\
    \hline
    
    berechne & calculate (Imp.) & \textipa{b\textschwa\textscr\textepsilon xne\textlengthmark} & \textipa{b\textschwa\textscr\textepsilon xn\textschwa} \\
    
    bidde & please & \textipa{b\textsci tde\textlengthmark} & \textipa{b\textsci d\textschwa}\\
    
    Midde & middle & \textipa{m\textsci tde\textlengthmark} & \textipa{m\textsci d\textschwa}\\
    
    odder & or & \textipa{\textopeno td\textschwa \textscr} & \textipa{\textopeno d\textturna}\\ 
    
    buuch & book (Imp.) & \textipa{bu\textlengthmark u\textlengthmark x} & \textipa{bu\textlengthmark x}\\

    ääner & one & \textipa{\ae\textlengthmark\ae\textlengthmark n\textschwa\textscr} & \textipa{\ae\textlengthmark n\textturna}\\

    òòmens & in the evening & \textipa{o\textlengthmark ò\textlengthmark \`m\textschwa ns} & \textipa{\textopeno\textlengthmark mns} \\ 

    Schdigg & piece & \textipa{\textesh d\textsci\c{c}k} & \textipa{\textesh d\textsci g}\\
    
    \hline

\end{tabular}
\caption{Examples of G2P Errors. Corrections of transcriptions are estimates by a native speaker.}
\label{tab:g2p_errors}
\end{table*}

Beyond general data limitations, several issues emerged specifically during the G2P conversion process. Representative examples for each of the following issues can be found in Table \ref{tab:g2p_errors}.

Firstly, the most frequent detected IPA symbol in the output is the length mark /\textipa{\textlengthmark}/, indicating that the model predicts a high number of lengthened vowels. However, through manual inspection, it becomes clear that vowel length is often assigned incorrectly. Words ending in the letter “e”, are often transcribed to end in /\textipa{e\textlengthmark}/, which is incorrect. This mistake stands out in particular as the correct transcription would generally be /\textipa{\textschwa}/, just like in Standard German. Also, it appears to misclassify double vowels, which are common in dialectal writing, as a sequence of two lengthened vowels.

Additionally, a common phenomenon in dialectal writing is the occurrence of a doubling of the letter “d”, which is rather rare in Standard German and instead often corresponds to the Standard German “tt”. This double consonant is misinterpreted by the model as /\textipa{td}/ on several occasions, which reflects an incorrect segmentation.

Lastly, the Epitran~\citep{Mortensen-et-al:2018} G2P model generated some characters which are impossible in the German phoneme inventory, such as /\textipa{\`m}/, showing expected issues with processing the special characters “á” and “ò”.

\section{Dialectal TTS Modeling}
The primary motivation behind creating Saar-Voice is to enable the development and evaluation of multi-speaker TTS systems. With recent advances in multilingual and multi-speaker (zero-shot) TTS models, such as XTTS~\citep{casanova24_interspeech} and ZMM-TTS~\citep{gong2023zmm}, it is now possible to investigate how well multilingually pretrained systems can generalize to closely related dialectal varieties, even when only limited data are available. 

Both models include German in their pretraining, making adaptation to Saarbrücken dialect potentially feasible. However, recent research on low-resource Bildts (a Dutch variety)~\citep{do25_blizzard} observed speaker–language entanglement when adapting ZMM-TTS for unseen speakers (zero-shot), and for seen speakers, there were also multiple cases of mispronunciation~\citep{alabi25_blizzard}. In contrast, ~\citet{pine25_blizzard} shows that combining Dutch and Bildts and using the StyleTTS~\citep{NEURIPS2023_3eaad2a0} architecture is sufficient for the same task. Hence, our dataset will help to better understand how multilingually pretrained TTS models and architectures generalize to closely related dialectal varieties.

Such evaluation of zero-shot and fine-tuned TTS models using the dataset are a goal of future work.

\section{Conclusion}
This paper presented the creation of a multi-speaker corpus for a low-resource German dialect. The dataset was constructed using partially OCR-based text extraction and manual normalization, and speech was produced by a variety of speakers. The resulting corpus provides both textual and phonetic representation suitable for speech technology research, with the option to expand the quantity of speech data. The challenges faced in the creation of the corpus reflect broader issues in low-resource dialect modeling. While they may introduce variability, they capture what authentic dialect usage looks like in native speakers.

Empirical validation of the dataset through downstream experiments remains an important next step. TTS synthesis experiments using the corpus are left for future work. Through possible zero-shot inference, model fine-tuning and training approaches, further evidence of the dataset's utility for speech technology applications in this low-resource dialect setting may be provided.

Beyond the dataset creation and proposing use cases, we aim to emphasize the necessity of collaborating with the speaker community. Without direct engagement with the dialect writers' scene and the speakers, the amount of texts would have been dramatically reduced. The existence of this corpus is therefore closely tied to the active dialect-writing community.

In dialectal research and Natural Language Processing (NLP) as a whole, methodological decisions should not be guided solely by technical feasibility or research interest, but also by the expectations and preferences of the speaker community itself. Especially in low-resource and non-standard contexts, technological developments interacts directly with questions of representation, identity, language maintenance and culture.

Our questionnaire, following \citet{blaschke-etal-2024-dialect}, asked speakers directly whether they would welcome more digital applications supporting dialect spoken in the Saarland. The large majority of our participants expressed agreement (strong agreement: n=6; agreement: n=2; neutral: n=1). Similarly, most respondents agree that language technologies can contribute to dialect preservation (strong agreement: n=5; agreement: n=3; neutral: n=1). Despite the limited sample size and biased participant pool, these responses suggest that technological support may be perceived rather positively within the community.

Thus, we consider it an important detail that dialect NLP projects have a contact to the various communities using dialects for getting 
\begin{itemize}
  \item access to written texts and spoken materials,
  \item recommendations for book authors and relevant sources,
  \item insights into possible or prominent spelling patterns / conventions and variations,
  \item feedback on acceptable and desirable technological applications.
\end{itemize}

Community engagement does and should not only serve as a practical resource for data collection, but also as a means of aligning technological development with actual users.

\section{Acknowledgments}

We would like to thank all speakers who contributed their voices to this corpus, without whom this resource would not have been possible. Many thanks to the anonymous reviewers for their time and efforts, and for the extensive feedback and questions. Thank you also to our colleague Çağla Kints for carefully proof-reading this paper.
Jesujoba Alabi was funded by the Deutsche Forschungsgemeinschaft (DFG, German Research Foundation) – Project-ID 232722074 – SFB 1102.


\section{Supplementary Materials}



\subsection{Ethical Considerations}
\textbf{Informed consent.} Participants signed a consent from in which they agreed to the process of the recording as well as the anonymized storage, sharing and processing of the recorded data. No personally identifying information of speakers was recorded, and all data is exclusively linked to an anonymized, randomized speaker ID. Participants were also informed that they could withdraw their consent at any time.

\textbf{Data protection.} All recorded data is stored in an anonymized form, only linking to the randomized speaker ID, on secure university servers.

\textbf{Voluntariness.} Participation was completely voluntary and occurred without compensation. No material incentives were offered to participants, and participation had no academic or professional consequences in any way.

\textbf{Participant well-being.} Great care was taken of keeping the duration of recording sessions to around one hour, though participants were free to end their sessions early if they wished. Breaks were always possible, but taken at least after the recording of 100 sentences. Participants had the option to prepare for the recording, as they received the material to be read during their session digitally around one week in advance. 

\textbf{Community sensitivity.} The dataset represents just a subset of the regional dialects of the Saarland. The process of curating the data involved no evaluation of dialectal “correctness”, including no aim to standardize or change dialect varieties. The textual content of the dataset reflects the original authors' perspectives and was not modified or filtered to promote or follow any particular set of views.








\section{Bibliographical References}\label{sec:reference}
\bibliographystyle{lrec2026-natbib}
\bibliography{lrec2026-example}

\section{Language Resource References}
\label{lr:ref}
\bibliographystylelanguageresource{lrec2026-natbib}
\bibliographylanguageresource{languageresource}

\end{document}